\documentclass{article}

\usepackage[nonatbib,final]{neurips_2022}

\usepackage[utf8]{inputenc} 
\usepackage[T1]{fontenc}    
\usepackage{hyperref}       
\usepackage{url}            
\usepackage{booktabs}       
\usepackage{amsfonts}       
\usepackage{nicefrac}       
\usepackage[nopatch=eqnum]{microtype}      
\usepackage{xcolor}         

\usepackage{xspace}
\usepackage{colortbl}
\usepackage{subcaption}
\usepackage[pdftex]{graphicx}
\usepackage{amsmath,amsfonts,amsthm,amssymb}

\newcommand{\sys}{\textsc{RoTaR}\xspace}
\newcommand{\base}{\textsc{base}\xspace}

\newcommand{\tapasbase}{\textsc{Tapas}$_{\base}$\xspace}

\newcommand{\zui}[1]{\textcolor{blue}{Zui: #1}}

\title{RoTaR: Efficient Row-Based Table Representation Learning via Teacher-Student Training}

\author{%
  Zui Chen \\
  Tsinghua University\\
  Beijing, China \\
  \texttt{chenzui19@mails.tsinghua.edu.cn}%
  \And
  Lei Cao \\
  MIT/U of Arizona \\
  Massachusetts, USA \\
  \texttt{caolei@mit.edu}%
  \And
  Sam Madden \\
  MIT \\
  Massachusetts, USA \\
  \texttt{madden@csail.mit.edu}
}

\begin{document}

\maketitle

\begin{abstract}
    We propose \sys, a row-based table representation learning method, to address the efficiency and scalability issues faced by existing table representation learning methods. The key idea of \sys is to generate query-agnostic row representations that could be re-used via query-specific aggregation. In addition to the row-based architecture, we introduce several techniques: cell-aware position embedding, teacher-student training paradigm, and selective backward to improve the performance of \sys model.
\end{abstract}

\section{Introduction}
\label{sec:intro}

Tabular data is one of the most widely used media for storing information. Table representation learning has a wide range of downstream applications such as table question answering, table search, table type detection, etc. It is thus vital to design an effective and practical solution for table representation learning. However, despite its popularity and importance in modern data science, table representation learning is not well addressed, compared to images, texts, or other media.

Most of the previous attempts \cite{TaBERT,TAPEX,TAPAS-1,TAPAS-2,RPT,UnifiedSKG,TabNet} apply recent progress in natural language processing (NLP), i.e., transformers and large language models (LMs). These works directly serialize the entire table together with a query or related utterance into a sequence as the input to an LM, which is pretrained on a sufficient amount of table corpus. 
However, as the most common and best practices for table representation learning, this approach suffers from scalability and efficiency issues. 

First, serializing a large table containing a large number of rows will result in a long sequence which is hard to process by classical transformer-based models, because the complexity of such models is quadratic to the length of the input sequence. To solve this problem, some works optimize the transformer structure \cite{ET-Survey}, while others use table specific solutions to reduce the complexity of attention computation, such as restricting attention computation to the same row or column \cite{MATE, TABBIE} or only between the schema and values \cite{TURL}. 
However, these approaches do not eliminate the scalability issue, because a pretrained LM is subject to a max sequence length constraint. For example, GPT-3 limits the input length to 2048 tokens, while BERT sets this limit as 512. 
A table with a small number of rows can easily exceeds it, causing inevitable truncation and thus loss of information.

Second, the serialization process takes the given query as input, leading to query-specific encoding. Because in real-world scenarios many queries concentrate on a few tables, repeatedly computing the table representation for every new incoming query is inefficient. 

To address the above problems, we proposed \sys which learns a {\it query-agnostic} {\it row-based} table representation. Rather than serialize the whole table, \sys takes each row as input and efficiently produces row level encodings which can be re-used by any queries. It then uses query-specific aggregation to produce the table representation on top of these row encodings.

\section{\sys Methodology}\label{sec:method}

\subsection{Overall Architecture}\label{sec:architecture}

\begin{figure}
  \centering
  
  \includegraphics[width=1.0\linewidth]{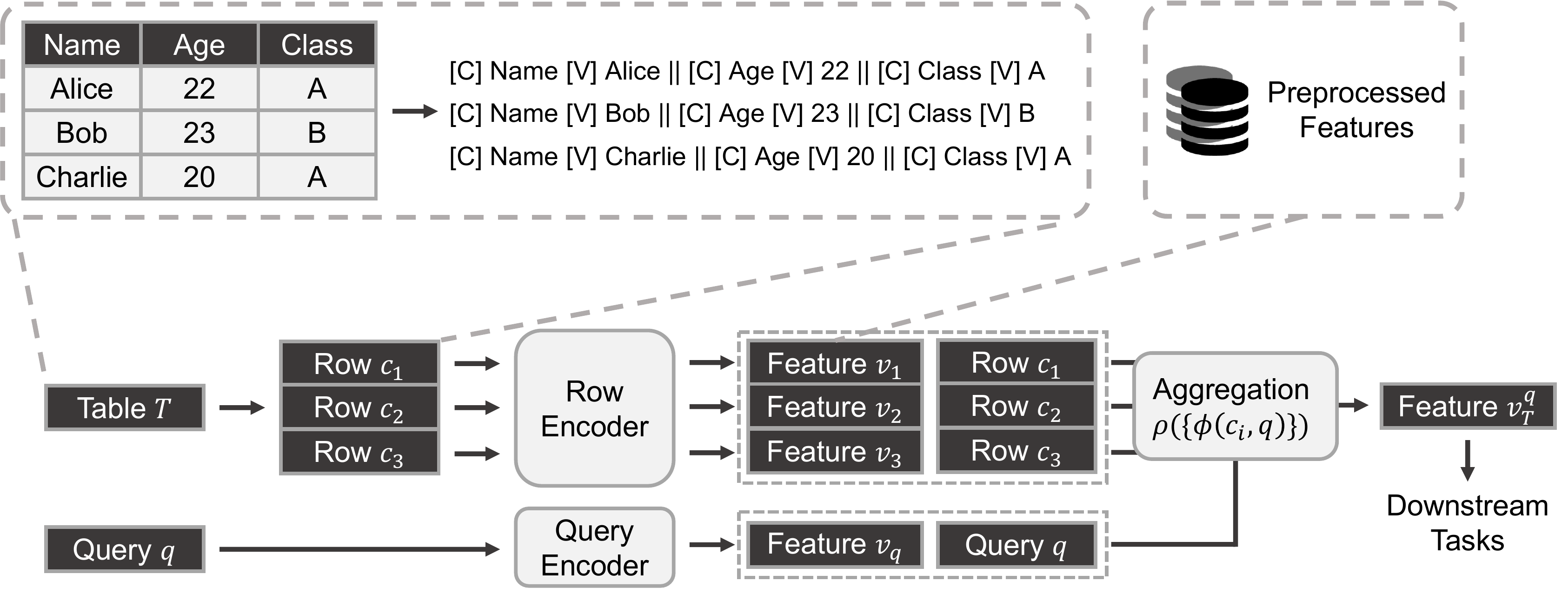}
  
  \caption{\sys model architecture.}
  \label{fig:Architecture}
  \vspace{-4mm}
\end{figure}

\noindent\textbf{Row Independence Observation.} 
Independencies in the table structure are the key to reduce the computational complexity of transformer-based models: irrelevant attention can be saved in transformer-based models without harming their performance. 
We observe that although the information in different rows should be aggregated to form the representation of the whole table, there is no strong correlation among different rows. 

Intuitively, a relational table can be viewed as a set of rows with homogeneous schema, because the order of the rows usually does not matter. The representation of a set is mathematically equivalent to an appropriate aggregation of the representations of each single item in the set \cite{Aggregation, DeepSets}. Therefore, table representation can be factorized into an aggregation of independent row representations.

Inspired by this observation, \sys uses a weight-shared row-based transformer model to encode each row in the table independently and ignores the inter-row correlation in this encoding (Fig.~\ref{fig:Architecture}). 

\sys consists of two components: query-agnostic row encoder $M$ and query-specific aggregation. After training on a dataset, the learned row representations by the row encoder can be preprocessed and stored. Therefore, answering an upcoming query only requires computing the aggregation. It does not have to repeatedly run the row encoder, thus saving a massive amount of time.

\textbf{Row Encoder.} More specifically, \sys considers every cell $T_{i,j}$ in a table $T$ as a textual cell, i.e., $T_{i,j} = T_{i,j;1}T_{i,j;2} \cdots T_{i,j;n}$ where each $T_{i,j;k}$ is a token. Similarly, each attribute in the schema $A$ is also viewed as textual, i.e., $A_j = A_{j;1}A_{j;2} \cdots A_{j;m}$, where each $A_{j;k}$ is a token. \sys thus serializes each cell $c_{i,j}$ by combining the attribute and the cell value $c_{i,j} = \texttt{[COL]} A_j \texttt{[VAL]} T_{i,j}$. Given a table with $N$ rows and $L$ columns, a shared encoder $M$ encodes each concatenated row $c_i = c_{i,1} \texttt{||} c_{i,2} \texttt{||} \cdots \texttt{||} c_{i,L}$ into a fixed-dimension vector $v_i = M(c_i)$. Notice that the obtained set of row representations $\lbrace v_1, v_2, \cdots, v_N \rbrace$ is query-agnostic. 

\textbf{Aggregation.} Then for each incoming query $q$, the resulted table representation can be computed by $v_T^q = \rho\left( \lbrace \phi(c_i, q) \rbrace_{i=1}^N \right)$, where $\phi$ is a learnable function that given a query $q$, extracts information from $c_i$. $\rho$ is an appropriate aggregation function \cite{Aggregation, DeepSets}. The function $\phi$ and $\rho$ together constitute a query-specific aggregation module. For instance, if a query $q$ is encoded as a vector $v_q$ of the same dimension with the row representations, setting $\phi(c_i,q) = v_i \odot v_q$ ($\odot$ stands for point-wise multiplication) and $\rho(X) = \frac{1}{\vert X \vert}\sum_{x \in X} x$ yields the table representation as the average row vector projected onto the query vector $v_q$.

\subsection{Query-agnostic Row Encoder}
\label{ssec:row}

The \sys model tackles the first fore-mentioned issue in scalability, i.e., encoding a table with a large number of rows. Because the transformer model is run separately for each row and the aggregation module does not have to use transformer-based models, \sys is capable of handling any tables with any number of rows. Note the number of columns is not of concern, because the number of columns is usually much smaller than the number of rows.

The design of the query-agnostic row encoder $M$ can be very flexible. For example, we can directly use a general-purpose pretrained LM like BERT \cite{BERT} or RoBERTa \cite{RoBERTa}, or pre-existing table representation models like TAPAS \cite{TAPAS-1, TAPAS-2} if we view each row as a table consisting of a single row. In addition to directly adopt the existing methods, we also propose two new techniques customized to row-based table representation learning.

\begin{figure}
  \centering
  \begin{subfigure}{0.55\linewidth}
    \centering
    \includegraphics[width=1.0\linewidth]{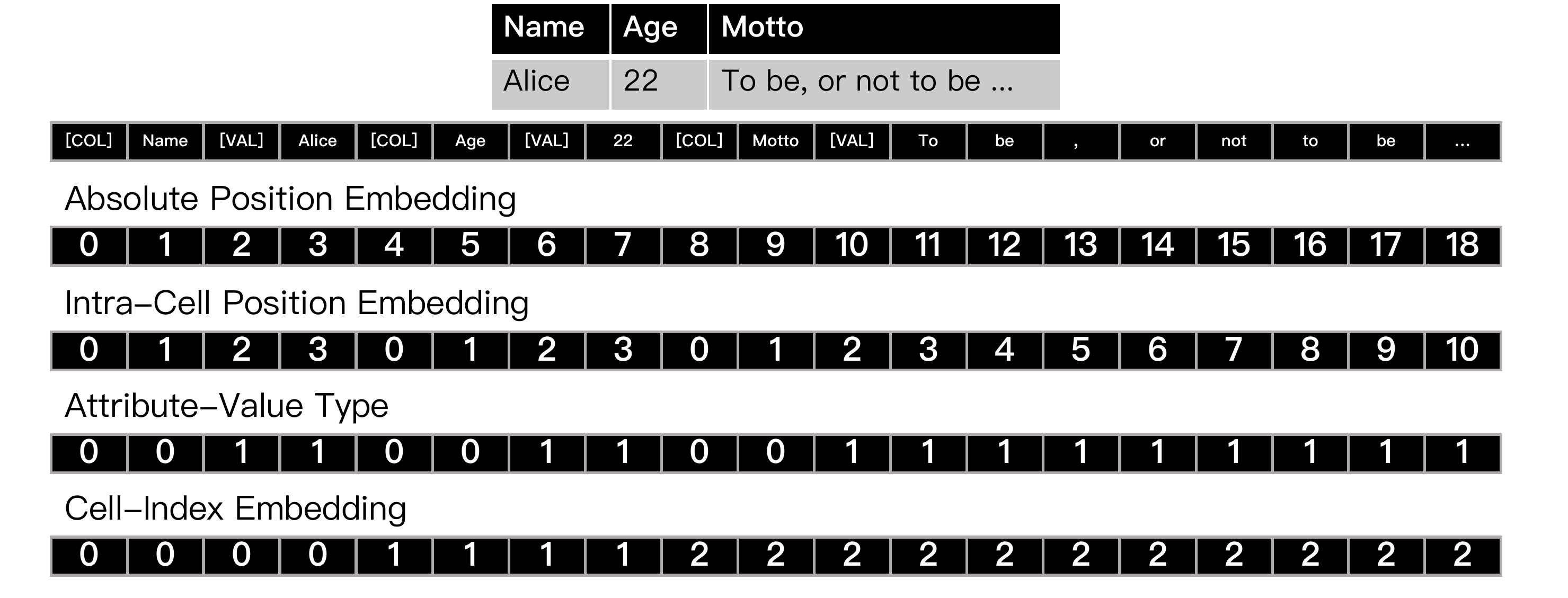}
  
    \subcaption[]{}
    \label{fig:Embedding}
  \end{subfigure}%
  \begin{subfigure}{0.45\linewidth}
    \centering
    \includegraphics[width=1.0\linewidth]{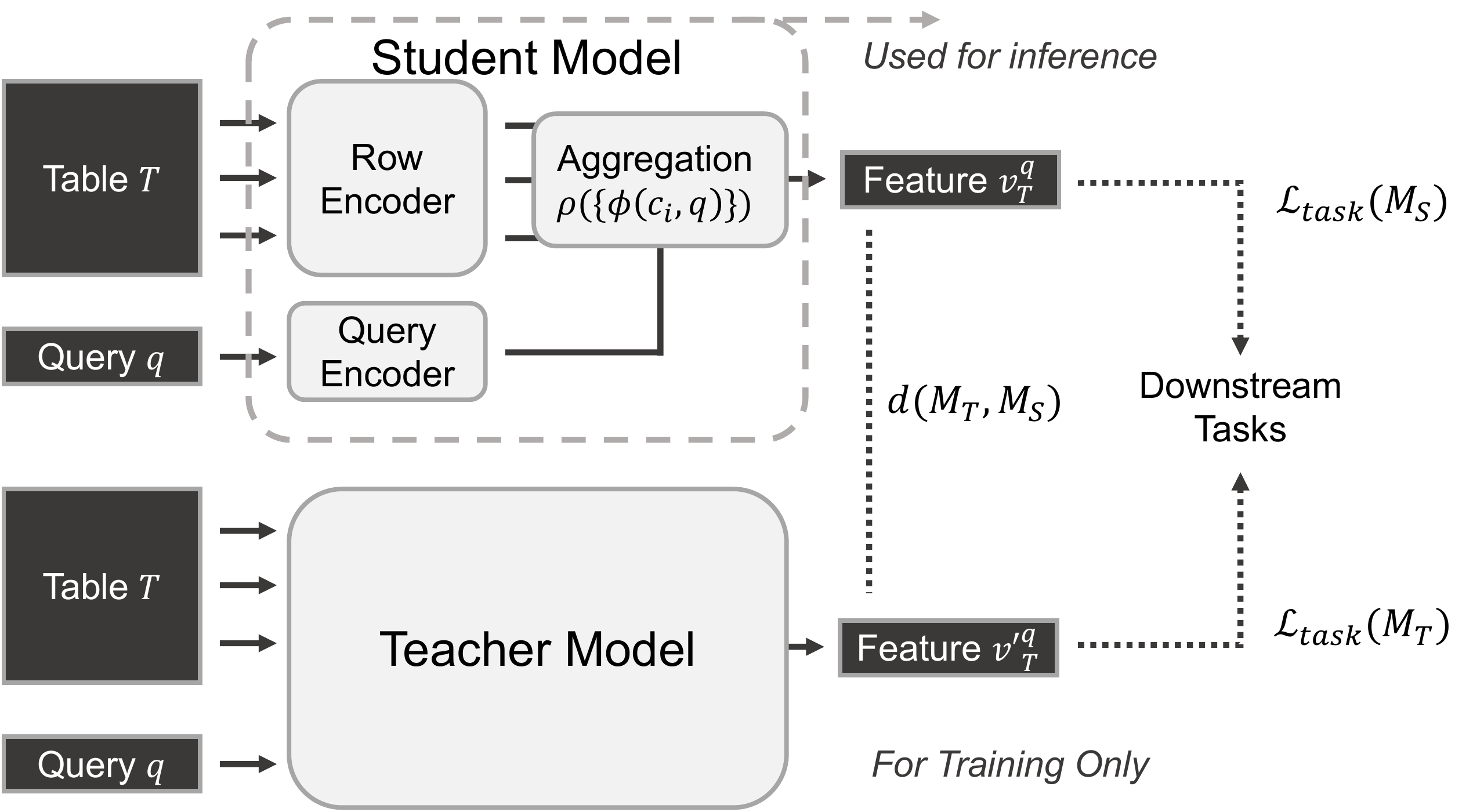}
  
    \subcaption[]{}
    \label{fig:Teacher-Student}
  \end{subfigure}
  \caption{(a). Cell-aware position embedding. (b). Teacher-student paradigm.}
  \vspace{-4mm}
\end{figure}

When using a general-purpose pretrained LM as the row encoder $M$, instead of directly feeding the serialized row $c_i = c_{i,1} \texttt{||} c_{i,2} \texttt{||} \cdots \texttt{||} c_{i,L}$ into the LM, we could take advantage of the structure of row to elaborate the position embedding design. Specifically, the position embedding $p^{T}_{i,j;k}$ of each token $T_{i,j;k}$ or $p^A_{j;k}$ of each token $A_{j;k}$ can be decomposed into inter-cell position embeddings (based on $j$ or $A_j$) and intra-cell position embeddings (based on $k$) \cite{dash-etal-2022-permutation}, which can then be customized for different purpose of use (Fig.~\ref{fig:Embedding}). For example, in many circumstances the order of attributes should be irrelevant to the representation of the row. By simply removing the absolute position embedding and the cell index embedding, the order of attributes is not perceived by the LM and thus the learned representation is robust against swapping order of columns.

\subsection{Query-specific Aggregation}
\label{ssec:aggregation}

The \sys model also tackles the fore-mentioned issue in efficiency, i.e., learning representation re-usable for different queries. While the learned row-based representation is query-agnostic, a query-specific aggregation module is introduced to produce query-specific table representation.

The design of query-specific adaption function $\phi$ can be very straightforward, for example, $\phi(v_i,q) = v_i \odot v_q$, or $\phi(v_i, q) = \texttt{MLP}(v_i \oplus v_q)$ ($\oplus$ stands for concatenation) or even $\phi(v_i, q) = \texttt{MLP}(v_i \oplus v_q \oplus \vert v_i-v_q \vert \oplus (v_i \odot v_q))$. Furthermore, notice that $\phi$ does not have to be differentiable or even numeric, traditional selective $\phi$ based on the textual input $c_i$ could also be used. For example, the n-gram similarity weighted embedding $\phi(c_i,q) = \texttt{ns}(c_i,q) \cdot v_i = \frac{2 \cdot \vert \texttt{n-gram}(c_i) \cap \texttt{n-gram}(q) \vert}{\vert \texttt{n-gram}(c_i) \vert + \vert \texttt{n-gram}(q) \vert} \cdot v_i$, or a hard threshold $\phi_{\alpha}(c_i,q) = [\texttt{ns}(c_i,q) > \alpha] \cdot v_i$.

The choice of aggregation function $\rho$ can be rather arbitrary, for example, $\texttt{Mean}$, $\texttt{Min}$, $\texttt{Max}$, $\texttt{LogSumExp}$, etc., are all feasible aggregation functions. However, the choice of aggregation naturally influences the table representation quality when handling different queries. Therefore, a learnable aggregation function $\rho$ could make the model more flexible. For instance, we could learn a multi-head projection aggregation function, which has a set of learnable parameters $\Theta = \lbrace \theta_l \rbrace_{l=1}^d$, and $\rho_\Theta(X) = \frac{1}{\vert X \vert} \sum_{x \in X} \texttt{Nonlinear}(x \odot \theta_l)$.

\subsection{Training Techniques}\label{ssec:training}

\subsubsection{Teacher-Student Paradigm}\label{sssec:teacher-student}
\sys simplifies the table representation process by ignoring inter-row interactions. Therefore, it pursues efficiency and scalability at the expense of inevitable but acceptable performance decay. However, with the help of the previous table representation methods during training time, the \sys model is able to improve its performance while still being efficient during inference time.

Consider the teacher-student paradigm which is widely used in model distillation \cite{TinyBERT,DistillBERT,MiniLM}. Training the student model to mimic the features generated by the highly performant teacher model can provide more differentiable information and boost the student model's learning process. The small and efficient student model is then used as a cost effective alternative to the original teacher model.

Technically, instead of only considering the loss function $\mathcal L = \mathcal L_{task}(M_S)$ of the downstream task during training, where $M_S$ is the student model, the loss in teacher-student paradigm uses $\mathcal L = \alpha \cdot \mathcal L_{task}(M_T) + \beta \cdot \mathcal L_{task}(M_S) + \gamma \cdot d(M_T,M_S)$, where $M_T$ is the teacher model, $d$ is the mean squared error distance between features or logits generated by the teacher and student model, and $\alpha, \beta, \gamma$ are tunable hyperparameters (Fig.~\ref{fig:Teacher-Student}). 

\subsubsection{Selective Backward}\label{sssec:selective-backward}

In practice, back-propagation through the aggregation of multiple rows could be unacceptable because of the GPU memory limit. However, since the row encoder $M$ is shared, \sys is able to only sample some rows to back-propagate. The sampling process could be done either randomly or weighted according to a traditional selective $\phi$, like n-gram similarity.

\section{Experiments}
\label{sec:exp}

We conduct preliminary experiments with the table fact verification task on the TabFact \cite{TabFact} dataset. In the table verification task, a query $q$ is a statement to be evaluated based on a table $T$. The TabFact dataset consists of $16\text{K}$ tables obtained from Wikipedia and $118\text{K}$ labeled statements. A common public data split is provided along with the dataset. The results are shown in Tab.~\ref{tab:Exp}.

\begin{table}
  \caption{Experiment result.}
  \label{tab:Exp}
  \centering
  \begin{tabular}{llc}
    \toprule
    Method                                          & Test Acc. ($\%$)                         & Inference Speed        \\
    \midrule
    {\color{gray} \tapasbase${}^\dagger$}           & {\color{gray} $78.5\% \pm 0.3\%$}        & {\color{gray} --}      \\
    \tapasbase (TabFact only)                       & $69.9\% \pm 3.8\%$                       & {\color{gray} --}      \\
    Table-BERT                                      & $65.1\%$                                 & {\color{gray} --}      \\
    \midrule
    \tapasbase (no query)                           & $50.3\%$                                 & $55\text{ms/table}$    \\
    \sys                                            & $63.6\%$                                 & $15\text{ms/table}$    \\
    \bottomrule
  \end{tabular}
\vspace{-4mm}
\end{table}

For fair comparison, we use the same model size to $\texttt{bert-base}$ and $\texttt{google/tapas-base}$. Further, because the \tapasbase best result${}^\dagger$ ($78.5\% \pm 0.3\%$) is pretrained on $6.2\text{M}$ table-text examples obtained from Wikipedia, we compare against the results reported by TAPAS using only TabFact data ($69.9\% \pm 3.8\%$) and Table-BERT ($65.1\%$). \footnote{The \tapasbase and \tapasbase (TabFact only) result is reported by \cite{TAPAS-2}, while the Table-BERT result is reported by \cite{TabFact}.}

Note since the \sys model prioritizes efficiency and scalability over performance, it slightly sacrifices performance in exchange for big speed up. The performance sacrifice mainly comes from the query-agnostic property instead of the row-independency
In particular, with a an accuracy drop of $\sim 6\%$, the \sys model with preprocessed feature vector is $\sim 3.7\text{x}$ faster than the TAPAS model, depending on the speed of the query encoder. 

Further, we also compare against a TAPAS model that same to \sys, is not aware of the queries beforehand. We then finetuned it under the same setting to \sys. This TAPAS model achieves an accuracy of $50.3\%$, which is much lower than the accuracy of our \sys ($63.6\%$). This confirms that \sys is indeed able to produce query-agnostic representation.  

\section{Conclusion}
\label{sec:conclusion}
We propose \sys which uses a shared row-encoder to generate query-agnostic row representations and learns instance optimized aggregation function to produce query-specific table representation. Preliminary experiments on TabFact confirm that \sys significantly improves the scalability and efficiency of table representation learning, with limited performance drop. 
In the further, we will continue to optimize the \sys model to improve the speed-accuracy trade-off.


\medskip

\bibliographystyle{plain}
\bibliography{ref}

\begin{thebibliography}{10}

\bibitem{TabNet}
Sercan~{\"{O}}. Arik and Tomas Pfister.
\newblock Tabnet: Attentive interpretable tabular learning.
\newblock In {\em {AAAI} 2021, {IAAI} 2021, {EAAI} 2021}. {AAAI} Press, 2021.

\bibitem{TabFact}
Wenhu Chen, Hongmin Wang, Jianshu Chen, Yunkai Zhang, Hong Wang, Shiyang Li,
  Xiyou Zhou, and William~Yang Wang.
\newblock Tabfact: {A} large-scale dataset for table-based fact verification.
\newblock In {\em {ICLR} 2020}. OpenReview.net, 2020.

\bibitem{dash-etal-2022-permutation}
Sarthak Dash, Sugato Bagchi, Nandana Mihindukulasooriya, and Alfio Gliozzo.
\newblock Permutation invariant strategy using transformer encoders for table
  understanding.
\newblock In {\em Findings of the Association for Computational Linguistics:
  NAACL 2022}. Association for Computational Linguistics, 2022.

\bibitem{TURL}
Xiang Deng, Huan Sun, Alyssa Lees, You Wu, and Cong Yu.
\newblock {TURL:} table understanding through representation learning.
\newblock {\em {SIGMOD} Rec.}, 2022.

\bibitem{BERT}
Jacob Devlin, Ming{-}Wei Chang, Kenton Lee, and Kristina Toutanova.
\newblock {BERT:} pre-training of deep bidirectional transformers for language
  understanding.
\newblock In Jill Burstein, Christy Doran, and Thamar Solorio, editors, {\em
  {NAACL-HLT} 2019}. Association for Computational Linguistics, 2019.

\bibitem{MATE}
Julian Eisenschlos, Maharshi Gor, Thomas M{\"{u}}ller, and William~W. Cohen.
\newblock {MATE:} multi-view attention for table transformer efficiency.
\newblock In Marie{-}Francine Moens, Xuanjing Huang, Lucia Specia, and
  Scott~Wen{-}tau Yih, editors, {\em {EMNLP} 2021}. Association for
  Computational Linguistics, 2021.

\bibitem{TAPAS-2}
Julian~Martin Eisenschlos, Syrine Krichene, and Thomas M{\"{u}}ller.
\newblock Understanding tables with intermediate pre-training.
\newblock In Trevor Cohn, Yulan He, and Yang Liu, editors, {\em Findings of the
  Association for Computational Linguistics: {EMNLP} 2020, Online Event, 16-20
  November 2020}, volume {EMNLP} 2020 of {\em Findings of {ACL}}, pages
  281--296. Association for Computational Linguistics, 2020.

\bibitem{TAPAS-1}
Jonathan Herzig, Pawel~Krzysztof Nowak, Thomas M{\"{u}}ller, Francesco
  Piccinno, and Julian~Martin Eisenschlos.
\newblock Tapas: Weakly supervised table parsing via pre-training.
\newblock In Dan Jurafsky, Joyce Chai, Natalie Schluter, and Joel~R. Tetreault,
  editors, {\em {ACL} 2020}. Association for Computational Linguistics, 2020.

\bibitem{TABBIE}
Hiroshi Iida, Dung Thai, Varun Manjunatha, and Mohit Iyyer.
\newblock {TABBIE:} pretrained representations of tabular data.
\newblock In Kristina Toutanova, Anna Rumshisky, Luke Zettlemoyer, Dilek
  Hakkani{-}T{\"{u}}r, Iz~Beltagy, Steven Bethard, Ryan Cotterell, Tanmoy
  Chakraborty, and Yichao Zhou, editors, {\em {NAACL-HLT} 2021}. Association
  for Computational Linguistics, 2021.

\bibitem{TinyBERT}
Xiaoqi Jiao, Yichun Yin, Lifeng Shang, Xin Jiang, Xiao Chen, Linlin Li, Fang
  Wang, and Qun Liu.
\newblock Tinybert: Distilling {BERT} for natural language understanding.
\newblock In Trevor Cohn, Yulan He, and Yang Liu, editors, {\em Findings of the
  Association for Computational Linguistics: {EMNLP} 2020, Online Event, 16-20
  November 2020}, volume {EMNLP} 2020 of {\em Findings of {ACL}}. Association
  for Computational Linguistics, 2020.

\bibitem{TAPEX}
Qian Liu, Bei Chen, Jiaqi Guo, Morteza Ziyadi, Zeqi Lin, Weizhu Chen, and
  Jian{-}Guang Lou.
\newblock {TAPEX:} table pre-training via learning a neural {SQL} executor.
\newblock In {\em {ICLR} 2022}. OpenReview.net, 2022.

\bibitem{RoBERTa}
Yinhan Liu, Myle Ott, Naman Goyal, Jingfei Du, Mandar Joshi, Danqi Chen, Omer
  Levy, Mike Lewis, Luke Zettlemoyer, and Veselin Stoyanov.
\newblock Roberta: {A} robustly optimized {BERT} pretraining approach.
\newblock {\em {CoRR}}, 2019.

\bibitem{DistillBERT}
Victor Sanh, Lysandre Debut, Julien Chaumond, and Thomas Wolf.
\newblock Distilbert, a distilled version of {BERT:} smaller, faster, cheaper
  and lighter.
\newblock {\em {CoRR}}, 2019.

\bibitem{RPT}
Nan Tang, Ju~Fan, Fangyi Li, Jianhong Tu, Xiaoyong Du, Guoliang Li, Samuel
  Madden, and Mourad Ouzzani.
\newblock {RPT:} relational pre-trained transformer is almost all you need
  towards democratizing data preparation.
\newblock {\em Proc. {VLDB} Endow.}, 2021.

\bibitem{ET-Survey}
Yi~Tay, Mostafa Dehghani, Dara Bahri, and Donald Metzler.
\newblock Efficient transformers: {A} survey.
\newblock {\em {CoRR}}, 2020.

\bibitem{Aggregation}
Edward Wagstaff, Fabian Fuchs, Martin Engelcke, Ingmar Posner, and Michael~A.
  Osborne.
\newblock On the limitations of representing functions on sets.
\newblock In Kamalika Chaudhuri and Ruslan Salakhutdinov, editors, {\em {ICML}
  2019}, Proceedings of Machine Learning Research. {PMLR}, 2019.

\bibitem{MiniLM}
Wenhui Wang, Furu Wei, Li~Dong, Hangbo Bao, Nan Yang, and Ming Zhou.
\newblock Minilm: Deep self-attention distillation for task-agnostic
  compression of pre-trained transformers.
\newblock In Hugo Larochelle, Marc'Aurelio Ranzato, Raia Hadsell,
  Maria{-}Florina Balcan, and Hsuan{-}Tien Lin, editors, {\em {NeurIPS} 2020},
  2020.

\bibitem{huggingface}
Thomas Wolf, Lysandre Debut, Victor Sanh, Julien Chaumond, Clement Delangue,
  Anthony Moi, Pierric Cistac, Tim Rault, Rémi Louf, Morgan Funtowicz, Joe
  Davison, Sam Shleifer, Patrick von Platen, Clara Ma, Yacine Jernite, Julien
  Plu, Canwen Xu, Teven~Le Scao, Sylvain Gugger, Mariama Drame, Quentin Lhoest,
  and Alexander~M. Rush.
\newblock Transformers: State-of-the-art natural language processing.
\newblock In {\em Proceedings of the 2020 Conference on Empirical Methods in
  Natural Language Processing: System Demonstrations}. Association for
  Computational Linguistics, 2020.

\bibitem{UnifiedSKG}
Tianbao Xie, Chen~Henry Wu, Peng Shi, Ruiqi Zhong, Torsten Scholak, Michihiro
  Yasunaga, Chien{-}Sheng Wu, Ming Zhong, Pengcheng Yin, Sida~I. Wang, Victor
  Zhong, Bailin Wang, Chengzu Li, Connor Boyle, Ansong Ni, Ziyu Yao,
  Dragomir~R. Radev, Caiming Xiong, Lingpeng Kong, Rui Zhang, Noah~A. Smith,
  Luke Zettlemoyer, and Tao Yu.
\newblock Unifiedskg: Unifying and multi-tasking structured knowledge grounding
  with text-to-text language models.
\newblock {\em {CoRR}}, 2022.

\bibitem{TaBERT}
Pengcheng Yin, Graham Neubig, Wen{-}tau Yih, and Sebastian Riedel.
\newblock Tabert: Pretraining for joint understanding of textual and tabular
  data.
\newblock In Dan Jurafsky, Joyce Chai, Natalie Schluter, and Joel~R. Tetreault,
  editors, {\em {ACL} 2020}. Association for Computational Linguistics, 2020.

\bibitem{DeepSets}
Manzil Zaheer, Satwik Kottur, Siamak Ravanbakhsh, Barnabas Poczos, Russ~R
  Salakhutdinov, and Alexander~J Smola.
\newblock Deep sets.
\newblock In I.~Guyon, U.~Von Luxburg, S.~Bengio, H.~Wallach, R.~Fergus,
  S.~Vishwanathan, and R.~Garnett, editors, {\em Advances in Neural Information
  Processing Systems}. Curran Associates, Inc., 2017.

\end{thebibliography}

\appendix

\section{Appendix: Experiment Settings}\label{sec:appendix}
We use the huggingface \cite{huggingface} implementation of transformer models including BERT and TAPAS. All models share the parameter of virtual batch size $64$, learning rate $2 \times 10^{-5}$, weight decay $10^{-5}$, the AdamW optimizer, and cosine annealing with $2$ warm-up epochs. The training process uses early stopping of patience $8$. All experiments are run in half-precision on a cloud server with a single NVIDIA Tesla V100 Tensor Core GPU.

All features are extended to the same dimension of $2048$ by a transformation module, which is a two-layer neural network with hidden size equal to the original feature dimension ($768$ in case of \base models), $\texttt{LeakyReLU}$ activation with negative slope $0.01$ and dropout probability $0.1$. All models share the same downstream binary classifier, which is a three-layer neural network with hidden size $2048$, $\texttt{LeakyReLU}$ activation with negative slope $0.01$ and dropout probability $0.1$. The query encoder is a separate transformer \base model.

\end{document}